\documentclass[10pt,twocolumn,letterpaper]{article}

\usepackage{cvpr}
\usepackage{times}
\usepackage{epsfig}
\usepackage{graphicx}
\usepackage{amsmath}
\usepackage{amssymb}
\usepackage{caption}
\usepackage{url}
\usepackage[breaklinks=true,bookmarks=false]{hyperref}

\cvprfinalcopy % *** Uncomment this line for the final submission

\def\cvprPaperID{108} % *** Enter the CVPR Paper ID here

\ifcvprfinal\fi

\begin{document}

%%%%%%%%% TITLE
\title{WiCV 2022: The Tenth Women In Computer Vision Workshop}

\author{
 Doris Antensteiner$^1$, Silvia Bucci$^2$, Arushi Goel$^3$, Marah Halawa$^4$, Niveditha Kalavakonda$^5$,\\ Tejaswi Kasarla$^6$, Miaomiao Liu$^7$, Nermin Samet$^8$, Ivaxi Sheth$^9$ \\\\ $^1$Austrian Institute of Technology, $^2$Polytechnic of Turin, $^3$University of Edinburgh,\\$^4$Technical University of Berlin, $^5$Univeristy of Washington,  $^6$University of Amsterdam, \\ $^7$Australian National University, $^8$Ecole des Ponts ParisTech, $^9$Mila-Quebec AI, ETS Montreal\\
 \tt\small wicvcvpr2022-organizers@googlegroups.com
}

\maketitle

%%%%%%%%% ABSTRACT
\begin{abstract}
\thispagestyle{empty}
In this paper, we present the details of Women in Computer Vision Workshop - WiCV 2022, organized alongside the hybrid CVPR 2022 in New Orleans, Louisiana.  It provides a voice to a minority (female) group in the computer vision community and focuses on increasing the visibility of these researchers, both in academia and industry. WiCV believes that such an event can play an important role in lowering the gender imbalance in the field of computer vision. WiCV is organized each year where it provides a)~opportunity for collaboration between researchers from minority groups, b)~mentorship to female junior researchers,  c)~financial support to presenters to overcome monetary burden and d)~large and diverse choice of role models, who can serve as examples to younger researchers at the beginning of their careers. In this paper, we present a report on the workshop program, trends over the past years, a summary of statistics regarding presenters, attendees, and sponsorship for the WiCV 2022 workshop.

\end{abstract}

\section{Introduction}
While excellent progress has been made in a wide variety of computer vision research areas in recent years, similar progress has not been made in the increase of diversity in the field and the inclusion of all members of the computer vision community. Despite the rapid expansion of our field, females still only account for a small percentage of the researchers in both academia and industry. Due to this, many female computer vision researchers can feel isolated in workspaces which remain unbalanced due to the lack of inclusion.

The Women in Computer Vision workshop is a gathering for both women and men working in computer vision. It aims to appeal to researchers at all levels, including established researchers in both industry and academia (e.g. faculty or postdocs), graduate students pursuing a Masters or PhD, as well as undergraduates interested in research.  This aims to raise the profile and visibility of female computer vision researchers at each of these levels, seeking to reach women from diverse backgrounds at universities and industry located all over the world.

There are three key objectives of the WiCV workshop.
The first to increase the WiCV network and promote interactions between members of this network, so that female students may learn from professionals who are able to share career advice and past experiences. A mentoring banquet is run alongside the workshop. This provides a casual environment where both junior and senior women in computer vision can meet, exchange ideas and even form mentoring or research relationships.

The workshop's second objective is to raise the visibility of women in computer vision. This is done at both the junior and senior levels. Several senior researchers are invited to give high quality keynote talks on their research, while junior researchers are invited to submit recently published or ongoing works with many of these being selected for oral or poster presentation through a peer review process. This allows junior female researchers to gain experience presenting their work in a professional yet supportive setting. We strive for diversity in both research topics and presenters' backgrounds. The workshop also includes a panel, where the topics of inclusion and diversity can be discussed between female colleagues.

Finally, the third objective is to offer junior female researchers the opportunity to attend a major computer vision conference which they otherwise may not have the means to attend. This is done through travel grants awarded to junior researchers who present their work in the workshop via a poster session. These travel grants allow the presenters to not only attend the WiCV workshop, but also the rest of the CVPR conference. 

\section{Workshop Program}
\label{program}
The workshop program consisted of 4 keynotes, 7 oral presentations, 34 poster presentations, a panel discussion, and a mentoring session. As with previous years, our keynote speakers were selected to have diversity among topic, background, whether they work in academia or industry, as well as their seniority. It is crucial to provide a diverse set of speakers so that junior researchers have many different potential role models who they can relate to in order to help them envision their own career paths.

The workshop schedule was as follows:
\begin{itemize}
\item Introduction
\item Invited Talk 1: Marina Marie-Claire Höhne (Technische Universität Berlin, Germany), \textit{Improving Explainable AI by using Bayesian Neural Networks}
\item Oral Session 1
\begin{itemize}
\item Jennifer Hobbs, \textit{Deep Density Estimation Based on Multi-Spectral Remote Sensing Data for In-Field Crop Yield Forecasting}
\item Ranya Almohsen , \textit{Generative Probabilistic Novelty Detection with Isometric Adversarial Autoencoders.}
\item Sarah A. Schneider , \textit{A Comparative Analysis in the Realm of Anomaly Detection}
\end{itemize}
\item Invited Talk 2: Tatiana Tommasi(Polytechnic University of Turin, Italy), \textit{Reliable 2D and 3D Models for Open World Applications}

\item Poster Session (in person)

\item Invited Talk 3: Michal Irani ( Weizmann Institute of Science, Israel), \textit{“Mind Reading”:  Self-supervised decoding of visual data from brain activity}
\item Oral Session 2
\begin{itemize}
\item Maxine A Perroni-Scharf, \textit{Material Swapping for 3D Scenes using a Learnt Material Similarity Measure}
\item Mengyuan Zhang , \textit{Enriched Robust Multi-View Kernel Subspace Clustering. Presenter}
\item Asra Aslam , \textit{Detecting Objects in Less Response Time for Processing Multimedia Events in Smart Cities}
\item  Sonam Gupta , \textit{RV-GAN: Recurrent GAN for Unconditional Video Generation}
\end{itemize}

\item Invited Talk 4: Angela Yao (School of Computing at the National University of Singapore), \textit{Capturing and Understanding 3D Hands in Action}
\item Panel Discussion
\item Closing Remarks
\item Mentoring Session and Dinner (in person)
\begin{itemize}
\item Speaker: 
Angela Dai (Technical University of Munich, Germany)
\item Speaker: Djamila Aouada (University of Luxembourg)
\end{itemize}
\end{itemize}

\subsection{Hybrid Setting}
This year, the organization has been slightly modified as CVPR 2022 was held in hybrid setting. We had to make two plans, one for in-person attendance and one for virtual attendance. We made sure to make the virtual WiCV workshop as engaging and interactive as possible. Talks, oral sessions, and the panel was shared via zoom for the virtual attendances. While the poster session was repeated virtually a week after the conference, similar to the main conference setting. We also provided online mentoring sessions held via Zoom, for mentors and mentees that are able to attend only virtually. 

\section{Workshop Statistics}

Originally, the first workshop for WiCV was held in conjunction with CVPR 2015. Since then, the participation rate and number quality of submissions to WiCV have been steadily increasing. 
Following the examples from the editions held in previous years \cite{Akata18,Amerini19,Demir18,doughty2021wicv,goel2022wicv}, we were encouraged to collect the top quality submissions into workshop proceedings. By providing oral and poster presenters with the opportunity to publish their work in the conference's proceedings, we believe that the visibility of female researchers will be further increased.
This year, the workshop was held as a half day hybrid gathering, the virtual setting was over gather-town and Zoom, and the in-person setting was in New Orleans, Louisiana. Senior and junior researchers were invited to present their work, and poster presentations are included as already described in the previous Section \ref{program}.\\

The organizers for this year WiCV workshop are working in both academia and industry from various institutions located in different time zones. Their miscellaneous backgrounds and research areas have pledged the organizing committee a diverse perspective. Their research interests in computer vision and machine learning include video understanding, representation learning, 3D reconstruction, domain adaptation, domain generalization, vision and language, and semi/self-supervised learning in different application areas such as vision for robotics, and healthcare.

\begin{figure}[h]
\centering
\includegraphics[width=1\linewidth]{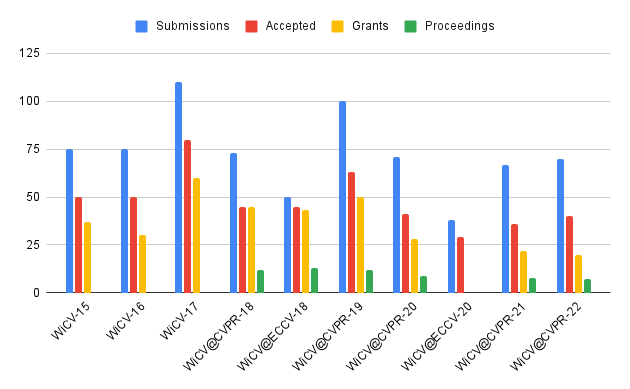}
\captionof{figure}{\textbf{WiCV Submissions.} The number of submissions over the past years of WiCV.}
\label{fig:sub}
\end{figure}

This year we had 70 high quality submissions from a wide range of topics and institutions. It is on par with WiCV@CVPR21. The most popular topics were deep learning architectures and techniques followed by video action and event recognition , segmentation and shape analysis, and medical application. 
Over all 70 submissions, 64 went into the review process. 7 papers were selected to be presented as oral talks and appeared into the CVPR22 workshop's proceedings, and 34 papers were selected to be presented as posters. Within the accepted submissions. The comparison with previous years is presented in Figure~\ref{fig:sub}. With the great effort of an interdisciplinary program committee consisting of 41 reviewers, the submitted papers were evaluated and received valuable feedback.

This year we kept WiCV tradition of last year's workshops \cite{Akata18,Amerini19,Demir18,doughty2021wicv, goel2022wicv} in providing grants to help the authors of accepted submissions participate in the workshop. The grants covered the conference registration fees itinerary (two ways flight), and two days accommodation for all the authors of accepted submissions who requested funding. 

\begin{figure}
\centering
\includegraphics[width=1\linewidth]{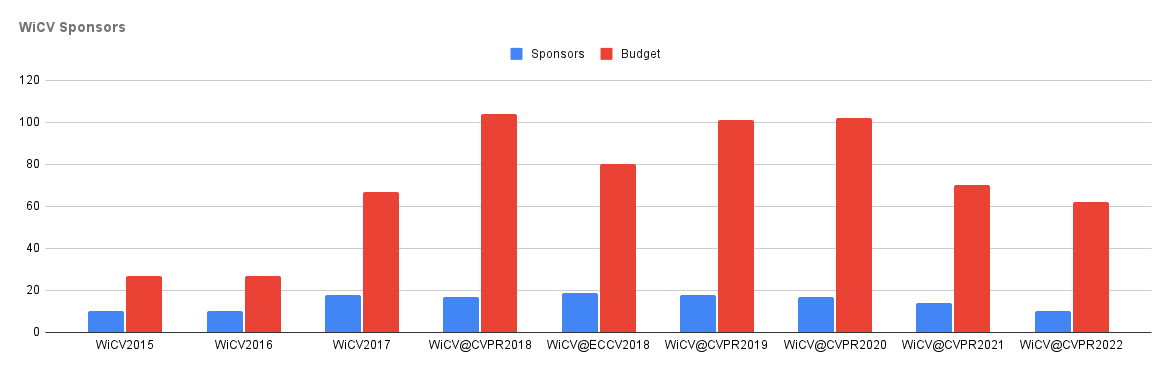}
\captionof{figure}{\textbf{WiCV Sponsors.} The number of sponsors and the amount of sponsorship for WiCV. The amount is expressed in US dollar (USD).}
\label{fig:spo}
\end{figure}

The total amount of sponsorship this year is \$62,000 USD with 10 sponsors, reaching a very good target. In Figure~\ref{fig:spo} you can find the details with respect to the past years. 

\section{Conclusions}
WiCV at CVPR 2022 has continued to be a valuable opportunity for presenters, participants and organizers in providing a platform to bring the community together. It continues to overcome the existing issue of gender balance prevailing around us and we hope that it has played an important part in making the community even stronger. It provided an opportunity for people to connect from all over the world from their personal comforts. With a high number of paper submissions and even higher number of attendees, we foresee that the workshop will continue the marked path of previous years and foster stronger community building with increased visibility, providing support, and encouragement inclusively for all the female researchers in academia and in industry.

\section{Acknowledgments}
First of all, we would like to thank our sponsors. We are very grateful to our other Platinum sponsors: Toyota Research Institute, Google, and Apple. We would also like to thank our Gold sponsor: Microsoft, and DeepMind; Silver Sponsors: Meta, Disney Research, and Zalando ; Bronze sponsors: Meshcapade, and Nvidia. We would also like to thank San Francisco Study Center as our fiscal sponsor, which helped to process our sponsorships and travel awards. We would also like to thank and acknowledge the organizers of previous WiCV, without the information flow and support from the previous WiCV organizers, this WiCV would not have been possible. Finally, we would like to acknowledge the time and efforts of our program committee, authors, reviewers, submitters, and our prospective participants for being part of WiCV network community.

\section{Contact}
\noindent \textbf{Website}: \url{https://sites.google.com/view/wicvcvpr2022/home}\\
\textbf{E-mail}: wicvcvpr2021-organizers@googlegroups.com\\
\textbf{Facebook}: \url{https://www.facebook.com/WomenInComputerVision/}\\
\textbf{Twitter}: \url{https://twitter.com/wicvworkshop}\\
\textbf{Google group}: women-in-computer-vision@googlegroups.com \\

{\small
\bibliographystyle{ieee}
\bibliography{egbib}
}

\end{document}